\documentclass[11pt,letterpaper]{article}
\usepackage{naaclhlt2015}
\usepackage{times}
\usepackage{latexsym}
\usepackage{graphicx}
\usepackage{color}
\usepackage[hyphens]{url}

\title{SemEval-2015 Task 3: Answer Selection in Community Question Answering}

\author{{Preslav Nakov}\hspace*{2.5mm}
        {Llu\'{i}s M\`arquez}\hspace*{2.5mm}
        {Walid Magdy}\hspace*{2mm}
        {Alessandro Moschitti}
        \\
        ALT Research Group, Qatar Computing Research Institute\\\smallskip\\
        {\bf James Glass}\\
        MIT Computer Science and Artificial Intelligence Laboratory\\\smallskip\\
        {\bf Bilal Randeree}\\
        Qatar Living\\}
\date{}

\begin{document}
\maketitle
\begin{abstract}
Community Question Answering (cQA)  provides new interesting research directions to the traditional Question Answering (QA) field, e.g., the exploitation of the interaction between users and the structure of related posts. 
In this context, we organized SemEval-2015 Task 3 on \emph{Answer Selection in cQA}, which included two subtasks:  (a) classifying answers as \textit{good}, \textit{bad}, or \textit{potentially relevant} with respect to the question, and (b) answering a YES/NO question with \textit{yes}, \textit{no}, or \textit{unsure}, based on the list of all answers. We set subtask A for Arabic and English on two relatively different cQA domains, i.e., the Qatar Living website for English, and a Quran-related website for Arabic. We used crowdsourcing on Amazon Mechanical Turk to label a large English training dataset, which we released to the research community.
Thirteen teams participated in the challenge with a total of 61 submissions: 24 primary and 37 contrastive.
The best systems achieved an official score (macro-averaged F$_1$) of 57.19 and 63.7 for the English subtasks A and B, and 78.55 for the Arabic subtask A.
\end{abstract}

\section{Introduction}
\label{sec:intro}

Many social activities on the Web, e.g., in forums and social networks, are accomplished by means of the community Question Answering (cQA) paradigm. User interaction in this context is seldom moderated, is rather open, and thus has little restrictions, if any, on who can post and who can answer a question.

On the positive side, this means that one can freely ask a question and expect some good, honest answers. On the negative side, it takes efforts to go through all possible answers and to make sense of them. It is often the case that many answers are only loosely related to the actual question, and some even change the topic. It is also not unusual for a question to have hundreds of answers, the vast majority of which would not satisfy a user's information needs; thus, finding the desired information in a long list of answers might be very time-consuming.

In our SemEval-2015 Task 3, we proposed two subtasks. First, subtask A asks for identifying the posts in the answer thread that answer the question \emph{well} vs. those that can be \emph{potentially useful} to the user (e.g., because they can help educate him/her on the subject) vs.~those that are just \emph{bad or useless}. This subtask goes in the direction of automating the answer search problem that we discussed above, and we offered it in two languages: English and Arabic. Second, for the special case of YES/NO questions, we propose an extreme summarization exercise (subtask B), which aims to produce a simple YES/NO overall answer, considering all \emph{good} answers to the questions (according to subtask A).

For English, the two subtasks are built on a particular application scenario of cQA, based on the Qatar Living forum.\footnote{\texttt{http://www.qatarliving.com/forum/}} However, we decoupled the tasks from the Information Retrieval component in order to facilitate participation, and to focus on aspects that are relevant for the SemEval community, namely on learning the relationship between two pieces of text. 

Subtask A goes in the direction of passage reranking,
where automatic classifiers are normally applied to pairs of questions and answer passages to derive a relative order between passages, e.g., see \cite{abs-cs-0605035,Jeon:2005:FSQ:1099554.1099572,shen-lapata:2007:EMNLP-CoNLL2007,Moschitti_et_al:ACL2007,Surdeanu2008}.
In recent years, many advanced models have been developed for automating answer selection, producing a large body of work.\footnote{\url{aclweb.org/aclwiki/index.php?title=Question_Answering_(State_of_the_art)}} For instance,  \newcite{wang:2007} proposed a probabilistic quasi-synchronous grammar to learn syntactic transformations from the question to the candidate answers; \newcite{heilman:naacl:2010} used an algorithm based on Tree Edit Distance (TED) to learn tree transformations in pairs; \newcite{wang_manning:acl:2010} developed a probabilistic model to learn tree-edit operations on dependency parse trees; and \newcite{yao:naacl:2013} applied linear chain CRFs with features derived from TED to automatically learn associations between questions and candidate answers.
One interesting aspect of the above research is the need for syntactic structures;
this is also corroborated in
\cite{sigir12,severyn-moschitti:2013:EMNLP}.  Note that answer selection can use models for textual entailment, semantic similarity, and for natural language inference in general. 

For Arabic, we also made use of a real cQA portal, the Fatwa website,\footnote{\texttt{http://fatwa.islamweb.net/}} where questions about Islam are posed by regular users and are answered by knowledgeable scholars. For subtask A, we used a setup similar to that for English, but this time each question had \emph{exactly} one correct answer among the candidate answers (see Section~\ref{sec:datasets} for detail); we did not offer subtask B for Arabic. 

Overall for the task, we needed manual annotations in two different languages and for two domains. For English, we built the Qatar Living datasets as a joint effort between MIT and the Qatar Computing Research Institute, co-organizers of the task, using Amazon's Mechanical Turk to recruit human annotators. For Arabic, we built the dataset automatically from the data available in the Fatwa website, without the need for any manual annotation.
We made all datasets publicly available, i.e., also usable beyond SemEval.


Our SemEval task attracted 13 teams, who submitted a total of 61 runs. The participants mainly focused on defining new features that go beyond question-answer similarity, e.g., author- and user-based, and spent less time on the design of complex machine learning approaches. Indeed, most systems used multi-class classifiers such as MaxEnt and SVM, but some used regression. Overall, almost all submissions managed to outperform the baselines using the official F$_1$-based score. In particular, the best system can detect a correct answer with an accuracy of about 73\% in the English task and 83\% in the easier Arabic task. For the extreme summarization task, the best accuracy is 72\%. 

An interesting outcome of this task is that the Qatar Living company, a co-organizer of the challenge, is going to use the experience and the technology developed during the evaluation excercise to improve their products, e.g., the automatic search of comments useful to answer users' questions.

The remainder of the paper is organized as follows:
Section~\ref{sec:task} gives a detailed description of the task,
Section~\ref{sec:datasets} describes the datasets,
Section~\ref{sec:scoring} explains the scorer,
Section~\ref{sec:results} presents the participants and the evaluation results,
Section~\ref{sec:features} provides an overview of the various features and techniques used by the participating systems,
Section~\ref{sec:discuss} offers further discussion, and
finally, Section~\ref{sec:conclusion} concludes and points to possible directions for future work.

\section{Task Definition}
\label{sec:task}

We have two subtasks:

\begin{itemize}
\item {\bf Subtask A:}
            Given a question (short title + extended description), and several community answers,
            classify each of the answers as
            \begin{itemize}
                \item[(a)] definitely relevant (good),
                \item[(b)] potentially useful (potential), or
                \item[(c)] bad or irrelevant (bad, dialog, non-English, other).
            \end{itemize}
\item {\bf Subtask B:}
      Given a YES/NO question (short title + extended description), and a list of community answers, decide whether the global answer to the question should be \textit{yes}, \textit{no}, or \textit{unsure}, based on the individual good answers. This subtask is only available for English.
\end{itemize}

\begin{figure*}[tbh]
\mbox{\includegraphics [scale=0.325] {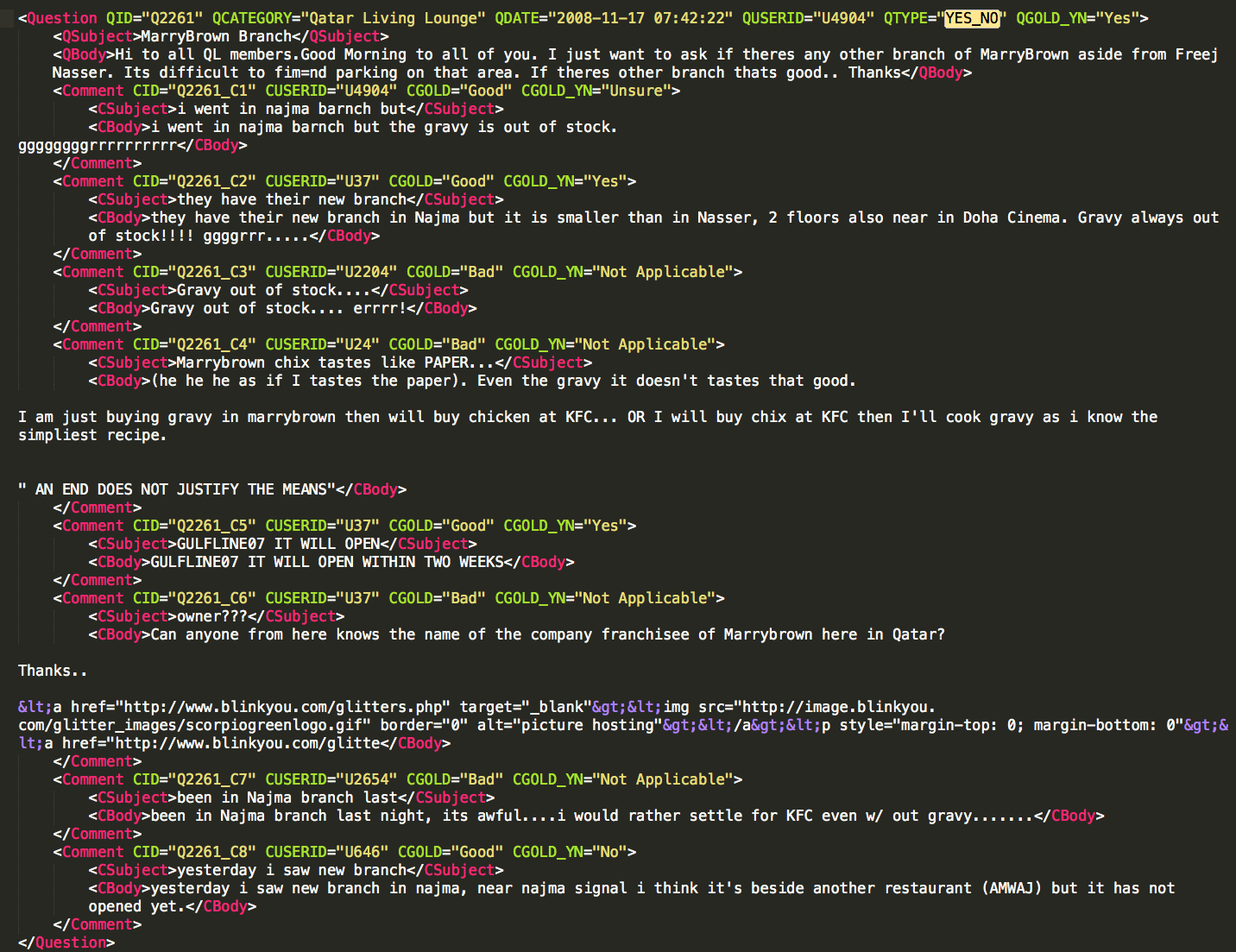}}
\caption{Annotated English question from the CQA-QL corpus.}
\label{fig:english:example}
\end{figure*}

\section{Datasets}
\label{sec:datasets}

We offer the task in two languages, English and Arabic, with some differences in the type of data provided.
For English, there is a question (short title + extended description) and a list of several community answers to that question.
For Arabic, there is a question and a set of possible answers, which include
(i)~a highly accurate answer,
(ii)~potentially useful answers from other questions, and
(iii)~answers to random questions.
The following subsections provide all the necessary details.

\subsection{English Data: CQA-QL corpus}

The source of the CQA-QL corpus is the Qatar Living forum.
A sample of questions and answer threads was
selected and then manually filtered and annotated with the categories defined in the task.

We provided a split in three datasets: training, development, and testing.
All datasets were XML-formated and the text was encoded in UTF-8.

A dataset file is a sequence of examples (questions),
where each question has a subject and a body (text),
as well as the following attributes:

\begin{itemize}
  \setlength{\itemsep}{1pt}
  \setlength{\parskip}{0pt}
  \setlength{\parsep}{0pt}
  \item QID: question identifier;
  \item QCATEGORY: the question category, according to the Qatar Living taxonomy;
  \item QDATE: date of posting;
  \item QUSERID: identifier of the user asking the question; 
  \item QTYPE: type of question (GENERAL or YES/NO);
  \item QGOLD\_YN: for YES/NO questions only, an overall \emph{Yes}/\emph{No}/\emph{Unsure} answer based on all comments.
\end{itemize}

Each question is followed by a list of comments (or answers).
A comment has a subject and a body (text),
as well as the following attributes:

\begin{itemize}
  \setlength{\itemsep}{1pt}
  \setlength{\parskip}{0pt}
  \setlength{\parsep}{0pt}
  \item CID: comment identifier;
  \item CUSERID: identifier of the user posting the comment;
  \item CGOLD: human assessment about whether the comment is
        \emph{Good}, \emph{Bad}, \emph{Potential}, \emph{Dialogue}, \emph{non-English}, or \emph{Other}.
  \item CGOLD\_YN: human assessment on whether the comment suggests a \emph{Yes}, a \emph{No}, or an \emph{Unsure} answer.
\end{itemize}

At test time, CGOLD, CGOLD\_YN, and QGOLD\_YN are hidden,
and systems are asked to predict CGOLD for subtask A,
and QGOLD\_YN for subtask B;
CGOLD\_YN is not to be predicted.

Figure~\ref{fig:english:example} shows a fully annotated English YES/NO question
from the CQA-QL corpus.
We can see that it is asked and answered in a very informal way
and that there are many typos, incorrect capitalization, punctuation, slang, elongations, etc.
Four of the comments are good answers to the question, and four are bad.
The bad answers are irrelevant with respect to the YES/NO answer to the question as a whole,
and thus their CGOLD\_YN label is \emph{Not Applicable}.
The remaining four good answers predict
\emph{Yes} twice, \emph{No} once, and \emph{Unsure} once;
as there are more \emph{Yes} answers than the two alternatives,
the overall QGOLD\_YN is \emph{Yes}.


\subsection{Annotating the CQA-QL corpus}

The manual annotation was a joint effort between MIT and the Qatar Computing Research Institute, co-organizers of the task.
After a first internal labeling of a trial dataset (50+50 questions) by several independent annotators,
we defined the annotation procedure and prepared detailed annotation guidelines.
We then used Amazon's Mechanical Turk to collect human annotations for a much larger dataset.
This involved the setup of three HITs:

\begin{itemize}
  \item \textbf{HIT 1:} Select appropriate example questions and classify them as GENERAL vs.~YES/NO (QCATEGORY);
  \item \textbf{HIT 2:} For GENERAL questions, annotate each comment as
       \emph{Good}, \emph{Bad}, \emph{Potential}, \emph{Dialogue}, \emph{non-English}, or \emph{Other} (CGOLD);
  \item  \textbf{HIT 3:} For YES/NO questions, annotate the comments as in HIT 2 (CGOLD),
        plus a label indicating whether the comment answers the question with
        a clear \emph{Yes}, a clear \emph{No}, or in an undefined way, i.e., as \emph{Unsure} (CGOLD\_YN).
\end{itemize}

For all HITs, we collected annotations from 3-5 annotators for each decision,
and we resolved discrepancies using majority voting. Ties led to the elimination of some comments and sometimes even of entire questions.

We assigned the \emph{Yes}/\emph{No}/\emph{Unsure} labels at the question level (QGOLD\_YN) automatically,
using the \emph{Yes}/\emph{No}/\emph{Unsure} labels at the comment level (CGOLD\_YN).
More precisely, we labeled a YES/NO question as \emph{Unsure},
unless there was a majority of \emph{Yes} or \emph{No} labels among the \emph{Yes}/\emph{No}/\emph{Unsure} labels
for the comments that are labeled as \emph{Good}, in which case we assigned the majority label.

Table~\ref{table:statistics:english} shows some statistics about the datasets.
We can see that the YES/NO questions are about 10\% of the questions.
This makes subtask B generally harder for machine learning, as there is much less training data.
We further see that on average, there are about 6 comments per question,
with the number varying widely from 1 to 143.
About half of the comments are \emph{Good},\
another 10\% are \emph{Potential}, and the rest are \emph{Bad}.
Note that for the purpose of classification, \emph{Bad} is in fact a heterogeneous class
that includes about 50\% \emph{Bad}, 50\% \emph{Dialogue},
and also a tiny fraction of \emph{non-English} and \emph{Other} comments. 
We released the fine grained labels to the task participants
as we thought that having information about the heterogeneous structure of \emph{Bad}
might be helpful for some systems.
About 40-50\% of the YES/NO annotations at the comment level (CGOLD\_YN) are \emph{Yes},
with the rest nearly equally split between \emph{No} and \emph{Unsure},
with \emph{No} slightly more frequent.
However, at the question level, the YES/NO annotations (QGOLD\_YN)
have more \emph{Unsure} than \emph{No}.
Overall, the label distribution in development and testing is similar to that in training
for the CGOLD values,
but there are somewhat larger differences for QGOLD\_YN.

We further released the raw text of all questions and of all comments from Qatar Living,
including more than 100 million word tokens,
which are useful for training word embeddings, topic models, etc.

\begin{table}[tbh]
\small
\begin{center}
\begin{tabular}{lrrr}
\bf Category & \bf Train & \bf Dev & \bf Test\\
\hline
\hline
\bf Questions & \bf 2,600 & \bf 300 & \bf 329 \\
\it -- GENERAL  & \it 2,376 & \it 266 & \it 304\\
\it -- YES/NO   & \it 224   & \it 34 & \it 25 \\
\hline
\hline
\bf Comments &  \bf 16,541 &  \bf 1,645 & \bf 1,976\\
\it -- min per question & \it 1 & \it 1 & \it 1\\
\it -- max per question & \it 143 & \it 32 & \it 66\\
\it -- avg per question & \it 6.36 & \it 5.48 & \it 6.01\\
\hline
\hline
\bf CGOLD values & \bf 16,541 &  \bf 1,645 & \bf 1,976\\
\it -- Good & \it 8,069 & \it 875 & \it 997\\
\it -- Potential & \it 1,659 & \it 187 & \it 167\\
\it -- Bad & \it 6,813 & \it 583 & \it 812\\
\hspace{1em} \tiny \it -- Bad & \tiny \it 2,981 & \tiny \it 269 & \tiny \it 362\\
\hspace{1em} \tiny \it -- Dialogue & \tiny \it 3,755 & \tiny \it 312 & \tiny \it 435\\
\hspace{1em} \tiny \it -- Not English & \tiny \it 74 & \tiny \it 2 & \tiny \it 15\\
\hspace{1em} \tiny \it -- Other & \tiny \it 3 & \tiny \it 0 & \tiny \it 0\\
\hline
\bf CGOLD\_YN values & \bf 795 & \bf 115 & \bf 111\\
\it -- Yes & \it 346 & \it 62 & -- \\ 
\it -- No & \it 236 & \it 32 & --\\ 
\it -- Unsure & \it 213 & \it 21 & -- \\ 
\hline
\bf QGOLD\_YN values & \bf 224 & \bf 34 & \bf 25\\
\it -- Yes & \it 87 & \it 16 & \it 15\\
\it -- No & \it 47 & \it 8 & \it 4\\
\it -- Unsure & \it 90 & \it 10 & \it 6\\
\hline
\hline
\end{tabular}
\caption{Statistics about the English data.}
\label{table:statistics:english}
\end{center}
\end{table}

\subsection{Arabic Data: Fatwa corpus}


\begin{figure*}[tbh]
\mbox{\includegraphics [scale=0.62] {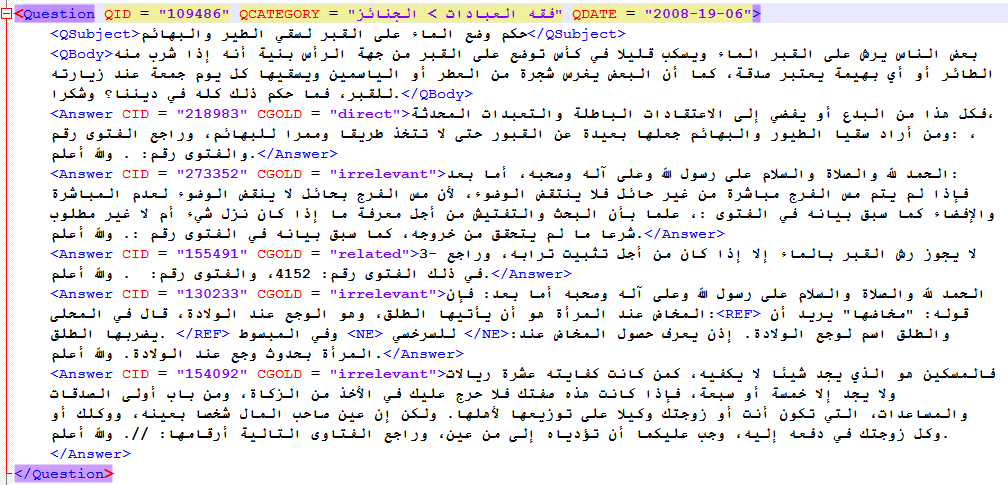}}
\caption{Annotated Arabic question from the Fatwa corpus.}
\label{fig:arabic:example}
\end{figure*}

For Arabic, we used data from the Fatwa website,
which deals with questions about Islam.
This website contains questions by ordinary users
and answers by knowledgeable scholars in Islamic studies.
The user question can be general, for example ``How to pray?'',
or it can be very personal, e.g., the user has a specific problem in his/her life
and wants to find out how to deal with it according to Islam.

Each question (Fatwa) is answered carefully by a knowledgeable scholar.
The answer is usually very descriptive:
it contains an introduction to the topic of the question,
then the general rules in Islam on the topic,
and finally an actual answer to the specific question and/or guidance on how to deal with the problem.
Typically, links to related questions are also provided to the user
to read more about similar situations and to look at related questions.

In the Arabic version of subtask A,
a question from the website is provided with a set of exactly five different answers.
Each answer of the provided five ones carries one of the following labels:

\begin{itemize}
  \setlength{\itemsep}{1pt}
  \setlength{\parskip}{0pt}
  \setlength{\parsep}{0pt}
  \item \emph{direct:} direct answer to the question;
  \item \emph{related:} not directly answering the question, but contains related information;
  \item \emph{irrelevant:} answer to another question not related to the topic.
\end{itemize}

Similarly to the English corpus, a dataset file is a sequence of examples (Questions),
where each question has a subject and a body (text),
as well as the following attributes:

\begin{itemize}
  \setlength{\itemsep}{1pt}
  \setlength{\parskip}{0pt}
  \setlength{\parsep}{0pt}
  \item QID: internal question identifier;
  \item QCATEGORY: question category;
  \item QDATE: date of posting.
\end{itemize}

Each question is followed by a list of possible answers.
An answer has a subject and a body (text),
as well as the following attributes:

\begin{itemize}
  \setlength{\itemsep}{1pt}
  \setlength{\parskip}{0pt}
  \setlength{\parsep}{0pt}
  \item CID: answer identifier;
  \item CGOLD: label of the answer, which is one of three: direct, related, or irrelevant.
\end{itemize}

Moreover, the answer body text can contain tags such as the following:

\begin{itemize}
  \setlength{\itemsep}{1pt}
  \setlength{\parskip}{0pt}
  \setlength{\parsep}{0pt}
  \item NE: named entities in the text, usually person names;
  \item Quran: verse from the Quran;
  \item Hadeeth: saying by the Islamic prophet.
\end{itemize}

Figure~\ref{fig:arabic:example} shows some fully annotated Arabic question from the Fatwa corpus.

\subsection{Annotating the Fatwa corpus}

\begin{table}[tbh]
\small
\begin{center}
\begin{tabular}{lrrrr}
\bf Category & \bf Train & \bf Dev & \bf Test & \bf Test30\\
\hline
\hline
\bf Questions & \bf 1,300 & \bf 200 & \bf 200 & \bf 30\\
\bf Answers &  \bf 6,500 &  \bf 1,000 & \bf 1,001 & \bf 151\\
\it -- Direct & \it 1,300 & \it 200 & \it 215 & \it 45\\
\it -- Related & \it 1,469 & \it 222 & \it 222 & \it 33\\
\it -- Irrelevant & \it 3,731 & \it 578 & \it 564 & \it 73\\
\hline
\end{tabular}
\caption{Statistics about the Arabic data.}
\label{table:statistics:arabic}
\end{center}
\end{table}


We selected the shortest questions and answers from IslamWeb to create our training, development and testing datasets. We avoided long questions and answers
since they are likely to be harder to parse, analyse, and classify. 
For each question, we labeled its answer as \emph{direct}, the answers of linked questions as \emph{related}, and we selected some random answers as \emph{irrelevant} to make the total number of provided answers per question equal to 5.

Table~\ref{table:statistics:arabic} shows some statistics about the resulting datasets. We can see that the number of \emph{direct} answers is the same as the number of questions, since each question has only one direct answer.

One issue with selecting random answers as \emph{irrelevant} is that the task is too easy;
thus, we manually annotated a special \emph{hard testset} of 30 questions (Test30), where we selected the \emph{irrelevant} answers using information retrieval to guarantee significant term overlap with the questions.
For the general testset, we used these 30 questions and 170 more where the \emph{irrelevant} answers were chosen randomly.



\section{Scoring}
\label{sec:scoring}

The official score for both subtasks is F$_1$, macro-averaged over the target categories:

\begin{itemize}
  \setlength{\itemsep}{1pt}
  \setlength{\parskip}{0pt}
  \setlength{\parsep}{0pt}
  \item For English, subtask A they are 
    \emph{Good}, \emph{Potential}, and \emph{Bad}.
  \item For Arabic, subtask A these are \emph{direct}, \emph{related}, and \emph{irrelevant}.
  \item For English, subtask B they are 
    \emph{Yes}, \emph{No}, and \emph{Unsure}.
\end{itemize}

We also report classification accuracy.

\begin{table}[tbh]
\small
\begin{center}
\begin{tabular}{@{}l@{ }@{ }l@{}}
\bf Team ID & \bf Affiliation and reference\\
\hline
Al-Bayan   & Alexandria University, Egypt \\
           & \cite{mohamed-EtAl:2015:SemEval} \\
CICBUAPnlp & Instituto Polit\'{e}cnico Nacional, Mexico\\
%
%
CoMiC      & University of T\"{u}bingen, Germany \\
           & \cite{rudzewitz-ziai:2015:SemEval} \\
ECNU       & East China Normal University, China \\
           & \cite{yi-wang-lan:2015:SemEval} \\
FBK-HLT    & Fondazione Bruno Kessler, Italy\\
           & \cite{vo-magnolini-popescu:2015:SemEval3} \\
HITSZ-ICRC & Harbin Institute of Technology, China\\
           & \cite{hou-EtAl:2015:SemEval1}\\
ICRC-HIT   & Harbin Institute of Technology, China\\
           & \cite{zhou-EtAl:2015:SemEval} \\
JAIST      & Japan Advance Institute of Science\\
           & and Technology, Japan \\
           & \cite{tran-EtAl:2015:SemEval} \\
QCRI       & Qatar Computing Research Institute, Qatar \\
           & \cite{nicosia-EtAl:2015:SemEval} \\
Shiraz     & Shiraz University, Iran \\
           & \cite{heydarialashty-EtAl:2015:SemEval} \\
VectorSLU  & MIT Computer Science and \\
           & Artificial Intelligence Lab, USA \\
           & \cite{belinkov-EtAl:2015:SemEval}\\
Voltron    & Sofia University, Bulgaria \\
           & \cite{zamanov-EtAl:2015:SemEval} \\
Yamraj     & Masaryk University, Czech Republic \\
\hline
\end{tabular}
\caption{The participating teams.}
\label{table:teams}
\end{center}
\end{table}

\section{Participants and Results}
\label{sec:results}

\begin{table}[t!]
\small
\begin{center}
\begin{tabular}{clrr}
& \bf Submission & \bf Macro F$_1$ & \bf \tiny Acc.\\
\hline
 & JAIST-contrastive1 & 57.29 & \tiny 72.67 \\
\bf 1 & \bf JAIST-primary & \bf 57.19 & \bf \tiny 72.52$_{1}$ \\
 & HITSZ-ICRC-contrastive1 & 56.44 & \tiny 69.43 \\
\bf 2 & \bf HITSZ-ICRC-primary & \bf 56.41 & \bf \tiny 68.67$_{5}$ \\
 & $^\star$QCRI-contrastive1 & 56.40 & \tiny 68.27 \\
 & HITSZ-ICRC-contrastive2 & 55.22 & \tiny 67.91 \\
 & ICRC-HIT-contrastive1 & 53.82 & \tiny 73.18 \\
\bf 3 & \bf $^\star$QCRI-primary & \bf 53.74 & \bf \tiny 70.50$_{3}$ \\
\bf 4 & \bf ECNU-primary & \bf 53.47 & \bf \tiny 70.55$_{2}$ \\
 & ECNU-contrastive1 & 52.55 & \tiny 69.48 \\
 & ECNU-contrastive2 & 52.27 & \tiny 69.38 \\
 & $^\star$QCRI-contrastive2 & 51.97 & \tiny 69.48 \\
\bf 5 & \bf ICRC-HIT-primary & \bf 49.60 & \bf \tiny 67.86$_{6}$ \\
 & $^\star$VectorSLU-contrastive1 & 49.54 & \tiny 70.45 \\
\bf 6 & \bf $^\star$VectorSLU-primary & \bf 49.10 & \bf \tiny 66.45$_{7}$ \\
\bf 7 & \bf Shiraz-primary & \bf 47.34 & \bf \tiny 56.83$_{9}$ \\
\bf 8 & \bf FBK-HLT-primary & \bf 47.32 & \bf \tiny 69.13$_{4}$ \\
 & JAIST-contrastive2 & 46.96 & \tiny 57.74 \\
\bf 9 & \bf Voltron-primary & \bf 46.07 & \bf \tiny 62.35$_{8}$ \\
 & Voltron-contrastive2 & 45.16 & \tiny 61.74 \\
 & Shiraz-contrastive1 & 45.03 & \tiny 62.55 \\
 & ICRC-HIT-contrastive2 & 40.54 & \tiny 60.12 \\
\bf 10 & \bf CICBUAPnlp-primary & \bf 40.40 & \bf \tiny 53.74$_{11}$ \\
 & CICBUAPnlp-contrastive1 & 39.53 & \tiny 52.33 \\
 & Shiraz-contrastive2 & 38.00 & \tiny 60.53 \\
\bf 11 & \bf Yamraj-primary & \bf 37.65 & \bf \tiny 45.50$_{12}$ \\
 & Yamraj-contrastive2 & 37.60 & \tiny 44.79 \\
 & Yamraj-contrastive1 & 36.30 & \tiny 39.57 \\
\bf 12 & \bf CoMiC-primary & \bf 30.63 & \bf \tiny 54.20$_{10}$ \\
 & CoMiC-contrastive1 & 23.35 & \tiny 50.56 \\
\hline
 & baseline: always ``Good'' & 22.36 & \tiny 50.46 \\ 
\end{tabular}
\caption{\textbf{Subtask A, English:} results for all submissions.
         The first column shows the rank for the primary submissions according to macro F$_1$,         
         and the subindex in the last column shows the rank based on accuracy.
         Teams marked with a $^\star$ include a task co-organizer.}
\label{table:results:subtaskA:english}
\end{center}
\end{table}

\begin{table}[tbh]
\small
\begin{center}
\begin{tabular}{clrr}
& \bf Submission & \bf Macro F$_1$ & \bf \tiny Acc.\\
\hline
1 & HITSZ-ICRC & 48.13 & \tiny 59.62$_{4}$\\
2 & $^\star$QCRI & 47.01 & \tiny 62.15$_{2}$\\
3 & ECNU & 46.57 & \tiny 61.34$_{3}$\\
4 & FBK-HLT & 42.61 & \tiny 62.40$_{1}$\\
5 & Shiraz & 40.06 & \tiny 48.53$_{10}$\\
6 & ICRC-HIT & 39.93 & \tiny 59.51$_{5}$\\
7 & $^\star$VectorSLU & 38.69 & \tiny 54.35$_{7}$\\
8 & CICBUAPnlp & 36.13 & \tiny 44.89$_{11}$\\
9 & JAIST & 35.09 & \tiny 54.61$_{6}$\\
10 & Voltron & 29.15 & \tiny 50.05$_{9}$\\
11 & Yamraj & 24.48 & \tiny 35.93$_{12}$\\
12 & CoMiC & 23.35 & \tiny 51.77$_{8}$\\
\hline
\end{tabular}
\caption{\textbf{Subtask A, English with \emph{Dialog} as a separate category:}
         results for the primary submissions.
         The first column shows the rank based on macro F$_1$,         
         the subindex in the last column shows the rank based on accuracy.
         Teams marked with a $^\star$ include a task co-organizer.}
\label{table:results:subtaskA:english:finegrained}
\end{center}
\end{table}

The list of all participating teams
can be found in Table~\ref{table:teams}.
The results for subtask A, English and Arabic,
are shown in Tables \ref{table:results:subtaskA:english}-\ref{table:results:subtaskA:english:finegrained} and \ref{table:results:subtaskA:arabic}-\ref{table:results:subtaskA:arabic30}, respectively;
those for subtask B are in Table~\ref{table:results:subtaskB:english}.
The systems are ranked by their macro-averaged F$_1$ scores for their primary runs (shown in the first column);
a ranking based on accuracy is also shown as a subindex in the last column.
We mark explicitly with an asterisk the teams that had a task co-organizer as a team member. This is for information only; these teams competed in the same conditions as everybody else.


\subsection{Subtask A, English}

Table~\ref{table:results:subtaskA:english} shows the results for subtask A, English,
which attracted 12 teams, which submitted 30 runs: 12 primary and 18 contrastive.
We can see that all submissions outperform,
in terms of macro F$_1$,
the majority class baseline that always predicts \emph{Good} (shown in the last line of the table);
for the primary submissions, this is so by a large margin.
However, in terms of accuracy, one of the primary submissions falls below the baseline;
this might be due to them optimizing for macro F$_1$ rather than for accuracy.

The best system for this subtask is JAIST,
which ranks first both in the official macro F$_1$ score (57.19) and in accuracy (72.52);
it used a supervised feature-rich approach,
which includes topic models and word vector representation,
with an SVM classifier.

The second best system is HITSZ-ICRC, which used an ensemble of classifiers. 
While it ranked second in terms of macro F$_1$ (56.41), it was only fifth on accuracy (68.67);
the second best in accuracy was ECNU, with 70.55.

The third best system, in both macro F$_1$ (53.74) and accuracy (70.50), is QCRI.
In addition to the features they used for Arabic (see the next subsection),
they further added cosine similarity based on word embeddings,
sentiment polarity lexicons, and metadata features such as the identity of the users asking and answering the questions or the existence of  acknowledgments.

Interestingly, the top two systems have contrastive runs
that scored higher than their primary runs both in terms of macro F$_1$ and accuracy,
even though these differences are small.
This is also true for QCRI's contrastive run in terms of macro F$_1$
but not in terms of accuracy,
which indicates that they optimized for macro F$_1$ for that contrastive run.
Note that ECNU was very close behind QCRI in macro F$_1$ (53.47),
and it slightly outperformed it in accuracy.

Note that while most systems trained a four-way classifier to distinguish \emph{Good}/\emph{Bad}/\emph{Potential}/\emph{Dialog}, where \emph{Bad} includes \emph{Bad}, \emph{Not English} and \emph{Other},
some systems targetted a three-way distinction \emph{Good}/\emph{Bad}/\emph{Potential}, following the grouping in Table~\ref{table:statistics:english},
as for the official scoring the scorer was merging \emph{Dialog} with \emph{Bad} anyway.

Table~\ref{table:results:subtaskA:english:finegrained} shows the results with four classes.
The last four systems did not predict \emph{Dialog},
and thus are severely penalized by macro F$_1$.
Comparing Tables \ref{table:results:subtaskA:english}
and \ref{table:results:subtaskA:english:finegrained},
we can see that the scores for the 4-way classification are up to 10 points 
lower than for the 3-way case.
Distinguishing \emph{Dialog} from \emph{Bad} turns out to be very hard:
e.g., HITSZ-ICRC achieved an F$_1$ of
76.52 for \emph{Good},
18.41 for \emph{Potential},
40.38 for \emph{Bad},
57.21 for \emph{Dialog};
however, merging \emph{Bad} and \emph{Dialog}
yielded an F$_1$ of 74.32 for the \emph{Bad+Dialog} category.
The other systems show a similar trend.

Finally, note that \emph{Potential} is by far the hardest class
(with an F$_1$ lower than 20 for all teams),
and it is also the smallest one, which amplifies its weight with F$_1$ macro;
thus, two teams (CoMiC and FBK-HLT) have chosen never to predict it.

\subsection{Subtask A, Arabic}

\begin{table}[tbh]
\small
\begin{center}
\begin{tabular}{clrr}
& \bf Submission & \bf Macro F$_1$ & \bf \tiny Acc.\\
\hline
\bf 1 & \bf $^\star$QCRI-primary & \bf 78.55 & \bf \tiny 83.02$_1$ \\
 & $^\star$QCRI-contrastive2 & 76.97 & \tiny 81.92 \\
 & $^\star$QCRI-contrastive1 & 76.60 & \tiny 81.82 \\
 & $^\star$VectorSLU-contrastive1 & 73.18 & \tiny 78.12 \\
\bf 2 & \bf $^\star$VectorSLU-primary & \bf 70.99 & \bf \tiny 76.32$_2$ \\
 & HITSZ-ICRC-contrastive1 & 68.36 & \tiny 73.93 \\
 & HITSZ-ICRC-contrastive2 & 67.98 & \tiny 73.23 \\
\bf 3 & \bf HITSZ-ICRC-primary & \bf 67.70 & \bf \tiny 74.53$_3$ \\
\bf 4 & \bf Al-Bayan-primary & \bf 67.65 & \bf \tiny 74.53$_3$ \\
 & Al-Bayan-contrastive2 & 65.70 & \tiny 72.53 \\
 & Al-Bayan-contrastive1 & 61.19 & \tiny 71.33 \\
 \hline
 & baseline: always ``irrelevant'' & 24.03 & \tiny 56.34 \\ 
\end{tabular}
\caption{\textbf{Subtask A, Arabic:} results for all submissions.
         The first column shows the rank for the primary submissions according to macro F$_1$,         
         and the subindex in the last column shows the rank based on accuracy.
         Teams marked with a $^\star$ include a task co-organizer.}
\label{table:results:subtaskA:arabic}
\end{center}
\end{table}

\begin{table}[tbh]
\small
\begin{center}
\begin{tabular}{clrr}
& \bf Submission & \bf Macro F$_1$ & \bf \tiny Acc.\\
\hline
\bf 1 & \bf $^\star$QCRI-primary & \bf 46.09 & \bf \tiny 48.34 \\
 & $^\star$QCRI-contrastive1 & 43.32 & \tiny 46.36 \\
 & $^\star$QCRI-contrastive2 & 43.08 & \tiny 49.67 \\
 & Al-Bayan-contrastive1 & 42.04 & \tiny 47.02 \\
 & HITSZ-ICRC-contrastive1 & 39.61 & \tiny 40.40 \\
 & HITSZ-ICRC-contrastive2 & 39.57 & \tiny 40.40 \\
\bf 2 & \bf HITSZ-ICRC-primary & \bf 38.58 & \bf \tiny 39.74 \\
 & $^\star$VectorSLU-contrastive1 & 36.43 & \tiny 43.05 \\
 \bf 3 & \bf $^\star$VectorSLU-primary & \bf 36.75 & \bf \tiny 37.09 \\
 \bf 4 & \bf Al-Bayan-primary & \bf 34.93 & \bf \tiny 38.41 \\
 & Al-Bayan-contrastive2 & 34.42 & \tiny 35.76 \\
 \hline
 & baseline: always ``irrelevant'' & 21.73 & \tiny 48.34 \\ 
\end{tabular}
\caption{\textbf{Subtask A, Arabic:} results for the 30 manually annotated Arabic questions.}
\label{table:results:subtaskA:arabic30}
\end{center}
\end{table}

Table~\ref{table:results:subtaskA:arabic} shows the results for subtask A, Arabic,
which attracted four teams, which submitted a total of 11 runs: 4 primary and 7 contrastive.
All teams performed well above a majority class baseline
that always predicts \emph{irrelevant}.

QCRI was a clear winner with a macro F$_1$ of 78.55 and accuracy of 83.02.
They used a set of features composed of lexical similarities
and word $[1,2]$-grams.  
Most importantly, they exploited the fact that there is at most one good answer for a given question:
they rank the answers by means of logistic regression, and label the top answer as \emph{direct}, the next one as \emph{related} and the remaining as \emph{irrelevant}
(a similar strategy is used by some other teams too).

Even though QCRI did not consider semantic models for this subtask,
and the second best team did, the distance between them is sizeable.

The second place went to VectorSLU (F$_1$=70.99, Acc=76.32),
whose feature vectors incorporated text-based similarities, embedded word vectors from both the question and answers, and features based on normalized ranking scores.  Their word embeddings were generated with word2vec \cite{word2vec}, and trained on the Arabic Gigaword corpus.  Their contrastive condition labeled the top scoring response as {\it direct}, the second best as {\it related}, and the others as {\it irrelevant}.  Their primary condition did not make use of this constraint.

Then come HITSZ-ICRC and Al-Bayan,
which are tied on accuracy (74.53),
and are almost tied on macro F$_1$: 67.70 vs. 67.65.
HITSZ-ICRC translated the Arabic to English and then extracted features
from both the Arabic original and from the English translation.
Al-Bayan had a knowledge-rich approach that used MADA for morphological analysis,
and then combined information retrieval scores with explicit semantic analysis
in a decision tree.

For all submitted runs, identifying the \emph{irrelevant} answers was easiest, with F$_1$ for this class ranging from 85\% to 91\%. This was expected, since most of these answers were randomly selected and thus the probability of finding common terms between them and the questions was low. The F$_1$ for detecting the \emph{direct} answers ranged from 67\% to 77\%, while for the \emph{related} answers, it was lowest: 47\% to 67\%.

Table~\ref{table:results:subtaskA:arabic30} presents the results for the 30 manually annotated Arabic questions, for which a search engine was used to find possibly \emph{irrelevant} answers.
We can see that the results are much lower than those reported in Table~\ref{table:results:subtaskA:arabic}, which shows that detecting \emph{direct} and \emph{related} answers is more challenging when the \emph{irrelevant} answers contain many common terms with the question.
The decrease in performance can be also explained by the different class distribution in training and testing, e.g., on the average, there are 1.5 \emph{direct} answers in Test30 vs. just 1 in \emph{training}, and the proportion of \emph{irrelevant} also changed (see Table~\ref{table:statistics:arabic}).
The team ranking changed too. QCRI remained the best-performing team, but the worst performing group now has one of its contrastive runs doing quite well. VectorSLU, which relies heavily on word overlap and similarity between the question and the answer experienced a relatively higher drop in performance compared to the rest.
In future work, we plan to study further the impact of selecting the \emph{irrelevant} answers in various challenging ways.

\subsection{Subtask B, English}

\begin{table}[tbh]
\small
\begin{center}
\begin{tabular}{clrr}
& \bf Submission & \bf Macro F$_1$ & \bf \tiny Acc.\\
\hline
\bf 1 & \bf $^\star$VectorSLU-primary & \bf 63.7 & \bf \tiny 72$_{1}$ \\
 & $^\star$VectorSLU-contrastive1 & 61.9 & \tiny 68 \\
\bf 2 & \bf ECNU-primary & \bf 55.8 & \bf \tiny 68$_{2}$ \\
 & ECNU-contrastive2 & 53.9 & \tiny 64 \\
\bf 3 & \bf $^\star$QCRI-primary & \bf 53.6 & \bf \tiny 64$_{3}$ \\
\bf 3 & \bf $^\diamond$HITSZ-ICRC-primary & \bf 53.6 & \bf \tiny 64$_{3}$ \\
 & ECNU-contrastive1 & 50.6 & \tiny 60 \\
 & $^\star$QCRI-contrastive2 & 49.0 & \tiny 56 \\
 & HITSZ-ICRC-contrastive1 & 42.5 & \tiny 60 \\
 & HITSZ-ICRC-contrastive2 & 42.4 & \tiny 60 \\
 & ICRC-HIT-contrastive2 & 40.3 & \tiny 60 \\
\bf 5 & \bf CICBUAPnlp-primary & \bf 38.8 & \bf \tiny 44$_{6}$ \\ 
 & ICRC-HIT-contrastive1 & 37.6 & \tiny 56 \\
\bf 6 & \bf ICRC-HIT-primary & \bf 30.9 & \bf \tiny 52$_{5}$ \\
\bf 7 & \bf Yamraj-primary & \bf 29.8 & \bf \tiny 28$_{8}$ \\
 & Yamraj-contrastive1 & 29.8 & \tiny 28 \\
 & CICBUAPnlp-contrastive1 & 29.1 & \tiny 40 \\
\bf 8 & \bf FBK-HLT-primary & \bf 27.8 & \bf \tiny 40$_{7}$ \\
 & $^\star$QCRI-contrastive1 & 25.2 & \tiny 56 \\
 & Yamraj-contrastive2 & 25.1 & \tiny 36 \\
 \hline
 & baseline: always ``Yes'' & 25.0 & \tiny 60 \\ 
\end{tabular}
\caption{\textbf{Subtask B, English:} results for all submissions.
         The first column shows the rank for the primary submissions according to macro F$_1$,         
         and the subindex in the last column shows the rank based on accuracy.
         Teams marked with a $^\star$ include a task co-organizer.
         The submission marked with a $^\diamond$ was corrected after the deadline.}
\label{table:results:subtaskB:english}
\end{center}
\end{table}

Table~\ref{table:results:subtaskB:english} shows the results for subtask B, English,
which attracted eight teams, who submitted a total of 20 runs: 8 primary and 12 contrastive.
As for subtask A, all submissions outperformed the majority class baseline
that always predicts \emph{Yes} (shown in the last line of the table).
However, this is so in terms of macro F$_1$ only;
in terms of accuracy, only half of the systems managed to beat the baseline.

For most teams, the features used for subtask B were almost the same as for subtask A,
with some teams adding extra features,
e.g., that look for positive, negative and uncertainty words
from small hand-crafted dictionaries.

Most teams designed systems that make Yes/No/Unsure decisions at the comment level,
predicting CGOLD\_YN labels
(typically, for the comments that were predicted to be \emph{Good}
by the team's system for subtask A),
and were then assigned a question-level label using majority
voting.\footnote{In fact, the authors of the third-best system HITSZ-ICRC
 submitted by mistake for their primary run
predictions for CGOLD\_YN instead of QGOLD\_YN;
the results reported in Table~\ref{table:results:subtaskB:english} for this team
were obtained by converting these predictions using simple majority voting.}
This is a reasonable strategy as it mirrors the human annotation process.
Some teams tried to extract features from the whole list of comments
and to predict QGOLD\_YN directly, but this yielded drop in performance.

The top-performing system, in both macro F$_1$ (63.7) and accuracy (72), is VectorSLU.
It is followed by ECNU with F$_1$=55.8, Acc=68.
The third place is shared by QCRI and HITSZ-ICRC,
which have exactly the same scores (F$_1$=53.6, Acc=64),
but different errors and different confusion matrices.
These four systems are much better than the rest; 
the next system is far behind at F$_1$=38.8, Acc=44.

Interestingly, once again there is a tie for the third place between the participating teams,
as was the case for subtask A, Arabic and English.
Note, however, that this time all top systems' primary runs performed better
than their corresponding contrastive runs, which was not the case for subtask A.

\section{Features and Techniques}
\label{sec:features}

Most systems were supervised,\footnote{The only two exceptions
were Yamraj (unsupervised) and CICBUAPnlp (semi-supervised).}
and thus the main efforts were focused on feature engineering.
We can group the features participants used into the following four categories:

\begin{itemize}
  \item \textbf{question-specific features:}
    e.g., length of the question,
    words/stems/lemmata/$n$-grams in the question, etc.
  \item \textbf{comment-specific features:}
    e.g., length of the comment,
    words/stems/lemmata/$n$-grams in the question,
    punctuation (e.g., does the comment contain a question mark),
    proportion of positive/negative sentiment words,
    rank of the comment in the list of comments,
    named entities (locations, organizations),
    formality of the language used,
    surface features (e.g., phones, URLs), etc.
  \item \textbf{features about the question-comment pair:}
    various kinds of similarity between the question and the comment
    (e.g., lexical based on cosine,
     or based on WordNet, language modeling, topic models such as LDA
     or explicit semantic analysis),
    word/lemma/stem/$n$-gram/POS overlap between the question and the comment
    (e.g., greedy string tiling, longest common subsequences, Jaccard coefficient, containment, etc.),
    information gain from the comment with respect to the question, etc.
  \item \textbf{metadata features:}
    ID of the user who asked the question,
    ID of the one who posted the comment,
    whether they are the same,
    known number of \emph{Good}/\emph{Bad}/\emph{Potential} comments
    (in the training data) written by the user who wrote the comment,
    timestamp,
    question category, etc.    
\end{itemize}

Note that the metadata features overlap with the other three groups
as a metadata feature is about the question, about the comment,
or about the question-comment pair.
Note also that the features above can be binary, integer, or real-valued,
e.g., can be calculated using various weighting schemes such as TF.IDF for words/lemmata/stems.

Although most participants focused on engineering features
to be used with a standard classifier such as SVM or a decision tree,
some also used more advanced techniques.
For example, some teams used sequence or partial tree kernels \cite{Moschitti:2006}.
Another popular technique was to use word embeddings,
e.g., modeled using convolution or recurrent neural networks,
or with latent semantic analysis, and also vectors trained using word2vec and GloVe \cite{pennington-socher-manning:2014:EMNLP2014},
as pre-trained on Google News or Wikipedia,
or trained on the provided Qatar Living data.
Less popular techniques included
dialog modeling for the list of comments for a given question,
e.g., using conditional random fields to model the sequence of comment labels
(\emph{Good}, \emph{Bad}, \emph{Potential}, \emph{Dialog}),
mapping the question and the comment to a graph structure and performing graph traversal,
using word alignments between the question and the comment,
time modeling,
and sentiment analysis.
Finally, for Arabic, some participants translated the Arabic data to English,
and then extracted features from both the Arabic and the English version;
this is helpful, as there are many more tools and resources for English than for Arabic.

When building their systems,
participants used a number of tools and resources
for preprocessing, feature extraction, and machine learning,
including
Deeplearning4J,
DKPro,
GATE,
GloVe,
Google translate,
HeidelTime,
LibLinear,
LibSVM,
MADA,
Mallet,
Meteor,
Networkx,
NLTK,
NRC-Canada sentiment lexicons,
PPDB,
sklearn,
Spam filtering corpus,
Stanford NLP toolkit,
TakeLab,
TiMBL,
UIMA,
Weka,
Wikipedia,
Wiktionary,
word2vec,
WordNet,
and WTMF.


There was also a rich variety of preprocessing techniques used,
including
sentence splitting,
tokenization,
stemming,
lemmatization,
morphological analysis (esp. for Arabic),
dependency parsing,
part of speech tagging,
temporal tagging,
named entity recognition,
gazetteer matching,
word alignment between the question and the comment,
word embedding,
spam filtering,
removing some content (e.g.,~all contents enclosed in HTML tags,
emoticons,
repetitive punctuation,
stopwords,
the ending signature,
URLs, etc.)
substituting (e.g.,~HTML character encodings and some common slang words),
etc.


\section{Discussion}
\label{sec:discuss}

The task attracted 13 teams and 61 submissions.
Naturally, the English subtasks were more popular
(with 12 and 8 teams for subtasks A and B, respectively; compared to just 4 for Arabic):
there are more tools and resources for English
as well as more general research interest.
Moreover, the English data followed the natural discussion threads in a forum,
while the Arabic data was somewhat artificial.

We have seen that all submissions managed to outperform,
on the official macro F$_1$ metric,\footnote{Curiously,
there was a close tie for the third place for all three subtask-language combinations.}
a majority class baseline for both subtasks and for both languages;
this improvement is smaller for English and much larger for Arabic.
However, if we consider accuracy, many systems fall below the baseline for English in both subtasks.

Overall, the results for Arabic are higher than those for English for subtask A, 
e.g., 
there is an absolute difference of over 21 points in macro F$_1$ (78.55 vs. 57.19)
for the top systems.
This suggests that the Arabic task was generally easier.
Indeed, it uses very formal polished language both for the questions and the answers
(as opposed to the noisy English forum data);
moreover, it is known a priori that each question can have at most one \emph{direct} answer,
and the teams have exploited this information.

However, looking at accuracy, the difference between the top systems for Arabic and English
is just 10 points (82.02 vs.~72.52).
This suggests that part of the bigger difference for F$_1$ macro comes from the measure itself.

Indeed, having a closer look at the distribution of the F$_1$ values for the different classes
before the macro averaging, we can see that the results are much more balanced for Arabic
(F$_1$ of 77.31/67.13/91.21 for \emph{direct/related/irrelevant}; with P and R very close to F$_1$)
than for English
(F$_1$ of 78.96/14.36/78.24 for \emph{Good/Potential/Bad}; with P and R very close to F$_1$).
We can see that the \emph{Potential} class is the hardest. This 
can hurt the accuracy but only slightly as this class is the smallest.
However, it can still have a major impact on macro-F$_1$ due to the effect of macro-averaging.

Overall, for both Arabic and English, it was much easier to 
recognize \emph{Good/direct} and \emph{Bad/irrelevant} examples (P, R, F$_1$ about 80-90),
and much harder to do so for \emph{Potential/related}
(P, R, F$_1$ around 67 for Arabic, and 14 for English).
This should not be surprising, as this intermediate category
is easily confusable with the other two:
for Arabic, these are answers to related questions,
while for English, this is a category that was quite hard for human annotators.



We should say that even though we had used majority voting to ensure agreement between
annotators, we were still worried about the the quality of human annotations
collected on Amazon's Mechanical Turk.
Thus, we asked eight people to do a manual re-annotation of the QGOLD\_YN labels for the test data.
We found a very high degree of agreement between each of the human annotators and the Turkers.
Originally, there were 29 YES/NO questions,
but we found that four of them were arguably general rather than YES/NO, and thus we excluded them.
For the remaining 25 questions, we had a discussion between our annotators
about any potential disagreement,
and finally, we arrived with a new annotation that changed the labels of three questions. 
This corresponds to an agreement of 22/25=0.88 between our consolidated annotation and the Turkers,
which is very high. 
This new annotation was the one we used for the final scoring.
Note that using the original Turkers' labels yielded slightly different scores
but exactly the same ranking for the systems.
The high agreement between our re-annotations and the Turkers
and the fact that the ranking did not change
makes us optimistic about the quality of the annotations for subtask A too
(even though we are aware of some errors and inconsistencies in the annotations).


\section{Conclusion and Future Work}
\label{sec:conclusion}

We have described a new task that entered SemEval-2015:
task 3 on Answer Selection in Community Question Answering.
The task has attracted a reasonably high number of submissions:
a total of 61 by 13 teams.
The teams experimented with a large number of features,
resources and approaches, and we believe that the lessons learned
will be useful for the overall development
of the field of community question answering.
Moreover, the datasets that we have created as part of the task,
and which we have released for use to the community,\footnote{\url{http://alt.qcri.org/semeval2015/task3/}}
should be useful beyond SemEval.

In our task description, we especially encouraged solutions going beyond simple keyword and bag-of-words matching, e.g., using semantic or complex linguistic information in order to reason about the relation between questions and answers. Although participants experimented with a broad variety of features (including semantic word-based representations, syntactic relations, contextual features, meta-information, and external resources), we feel that much more can be done in this direction. 
Ultimately, the question of whether complex linguistically-based representations and inference can be successfully applied to the very informal and ungrammatical text from cQA forums remains unanswered to a large extent. 

Complementary to the research direction presented by this year's task,
we plan to run a follow-up task at SemEval-2016, with a focus on answering \emph{new} questions,
i.e., that were not already answered in Qatar Living.
For Arabic, we plan to use a real community question answering dataset,
similar to Qatar Living for English.

\section*{Acknowledgments}

This research is developed by the Arabic Language Technologies (ALT) group at Qatar Computing Re- search Institute (QCRI) within the Qatar Foundation in collaboration with MIT. It is part of the Interactive sYstems for Answer Search (Iyas) project.


We would like to thank Nicole Schmidt from MIT for her help with setting up and running the Amazon Mechanical Turk annotation tasks.

\bibliographystyle{naaclhlt2015}
\bibliography{semeval2015_task3}

\end{document}